\pdfoutput=1

\documentclass[11pt]{article}

\usepackage{emnlp2021}

\usepackage{times}
\usepackage{latexsym}
\usepackage{booktabs}
\usepackage{times}
\usepackage{soul}
\usepackage{multirow}
\usepackage{tikz}
\usepackage{todonotes}
\usepackage{url}
\usepackage{graphicx}
\usepackage{amsmath}
\usepackage{amsthm}
\usepackage{bbm}
\usepackage{subcaption}
\usepackage{amsfonts}
\usepackage{amssymb}
\usepackage{algorithm}
\usepackage{algorithmic}

\usepackage[T1]{fontenc}

\usepackage[utf8]{inputenc}

\usepackage{microtype}

\newtheorem{theorem}{Theorem}

\newtheorem{assumption}{Assumption}

\title{Adapting by Pruning: 
A Case Study on BERT}

\author{Yang Gao, Nicolo Colombo \\
  Royal Holloway \\ University of London \\
  \texttt{\{yang.gao,nicolo.colombo\}} \\ 
  \texttt{@rhul.ac.uk} \\\And
  Wei Wang \\
  Beijing University of \\ Posts and Telecommunications \\
  \texttt{weiwang@bupt.edu.cn} 
  }

\begin{document}
\maketitle
\begin{abstract}
Adapting pre-trained neural models to downstream
tasks has become the standard practice
for obtaining high-quality models.
In this work, we propose a novel 
model adaptation paradigm, \emph{adapting by pruning},
which prunes neural connections 
in the pre-trained model to optimise the 
performance on the target task;
all remaining connections have their weights intact.
%
We formulate adapting-by-pruning as an 
optimisation problem with a differentiable loss and
propose an efficient algorithm to
prune the model. 
We prove that the algorithm is near-optimal
under standard assumptions and 
apply the algorithm to adapt BERT to 
some GLUE tasks.
Results suggest that our method can prune
up to 50\% weights in BERT while yielding
similar performance compared to the  
fine-tuned full model.
We also compare our method with other
state-of-the-art pruning methods and study the topological differences of their 
obtained sub-networks.
\end{abstract}

\section{Introduction}
\begin{figure*}
    \centering
    \includegraphics[width=0.9\textwidth]{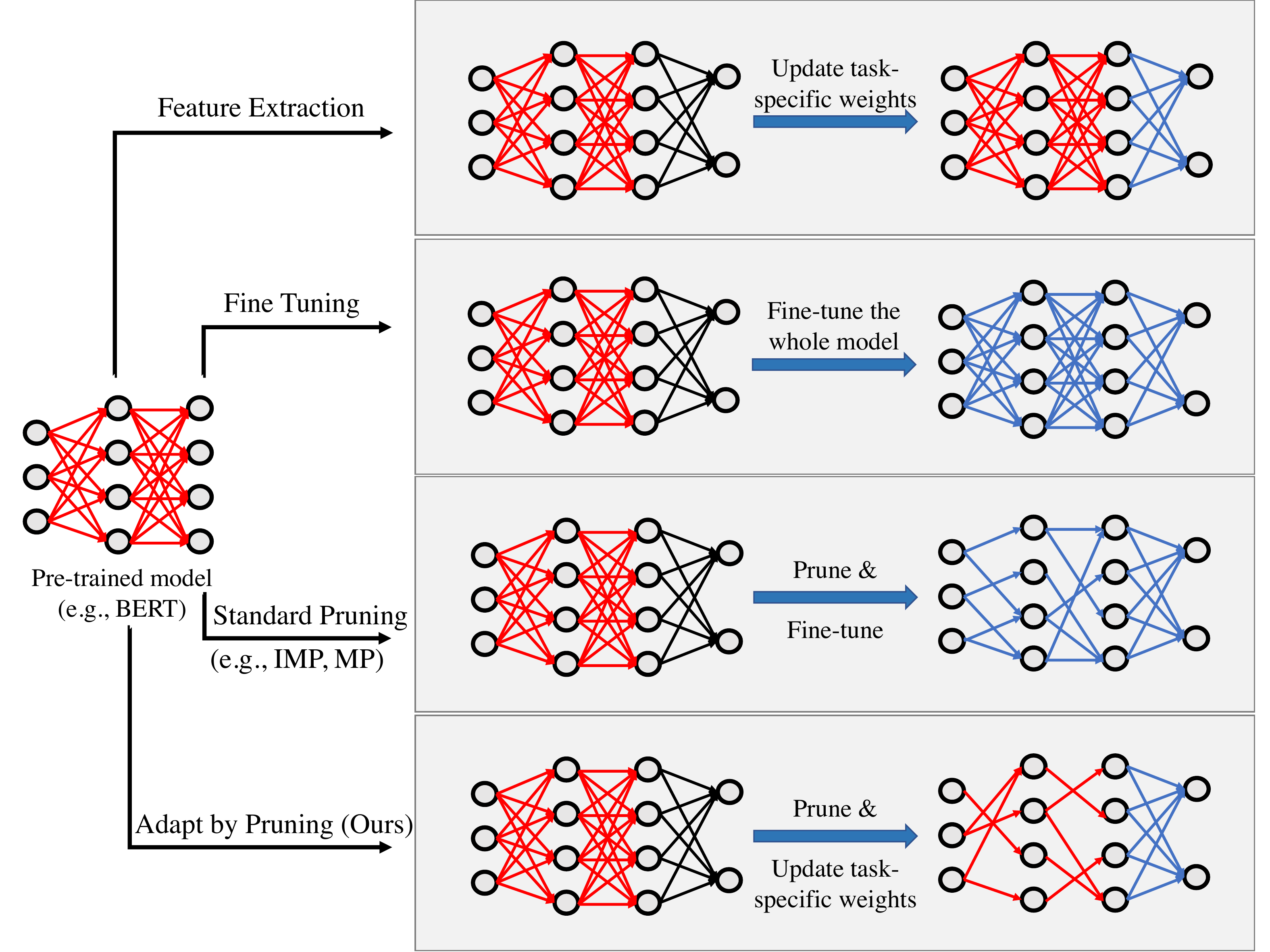}
    \caption{Comparison of different model adaptation paradigms. 
    %
    Red neural connections stand for the weights
    from the pre-trained model, blue for fine-tuned weights,
    and black for randomly initialised weights.
    }
    \label{fig:motivation}
\end{figure*}

From ResNet \cite{cvpr16-resnet} in computer vision to 
BERT \cite{naacl19-bert} in NLP,  large pre-trained networks
have become the standard starting point for developing 
neural models to solve downstream tasks.
The two predominant paradigms to adjust the pre-trained
models to downstream tasks are \emph{feature extraction}
and \emph{fine-tuning} \cite{rep4nlp-bert-transfer}.
Both paradigms require adding some task-specific layers
on top of the pre-trained model, but
in feature extraction, the weights of the pre-trained
model are ``frozen'' and only the task-specific layers are
trainable, while in fine-tuning, 
weights in both the pre-trained model
and the task-specific layers can be updated.
Variants of these two paradigms include
updating only selected layers 
in the original model \cite{arxiv20-finetune-bert}, and
the \emph{Adaptor} approach which
inserts trainable layers in between original layers \cite{nips17-adapter,icml19-adaptor}.
However, all these adaptation paradigms 
need to add new parameters and hence
\emph{enlarge the model size}.
Consequently, running the adapted models, 
even at inference time, 
requires considerable resources
and time,
especially on mobile devices
\cite{arxiv20-compress-bert-survey,arxiv-prune-survey}.
In this paper, we propose a novel
model adaptation paradigm that we call  \emph{adapt by pruning}.
Given a downstream task, we prune the 
\emph{task-irrelevant} neural connections
from the pre-trained model to optimise
the performance; 
all remaining connections 
inherit the corresponding weights
from the original pre-trained model.
Our paradigm is inspired by the observation
that large pre-trained models are highly
over-parameterised \cite{iclr19-lt} and 
pruning appropriate weights 
in large (randomly initialised)
neural models can yield strong performance
even \emph{without fine-tuning the weights}
\cite{nips19-supermask,nips20-supermask-sequence}.
%
%
Our paradigm is particularly suitable 
for the applications on mobile devices \cite{icml19-adaptor}:
the users only need to download the base model
once, and as new tasks arrive, 
they only need to download a task-specific binary masks 
to adapt the base model to the new tasks.
Table~\ref{table:best_of_both_world} compares
our method with multiple model adaptation 
and compression paradigms.
Fig.~\ref{fig:motivation} provides a graphical explanation of our method.

\begin{table}[]
    \centering
    \small
    \begin{tabular}{l | c c c c}
    \toprule
    & FullPerf & Eff & ReuseStruct & ReusePara \\
    \midrule
    Feat.Ext. & & & \checkmark & \checkmark \\
    FineTune & \checkmark & & \checkmark & \\
    Adaptor & \checkmark & & \checkmark & \checkmark \\
    Distillation & \checkmark & \checkmark &  &  \\
    Pruning & \checkmark & \checkmark & \checkmark & \\
    Ours & \checkmark & \checkmark & \checkmark & \checkmark\\
    \bottomrule
    \end{tabular}
    \caption{
    Our method can achieve near full model 
    performance (FullPerf) and 
    has high inference-time efficiency (Eff), 
    while reusing both the structure 
    (ReuseStruct) and weights (ReusePara) of the original pre-trained model. 
    }
    \label{table:best_of_both_world}
\end{table}

We exploit the idea that
pruning can be formulated as 
learning \emph{how to mask the pre-trained model} 
\cite{arxiv-our-binary}.
Let $f(x;w_0,p)$ be the 
neural model for a downstream task, where $x$
is the input, $w_0 \in \mathbb{R}^d$ are the 
weights of the pre-trained model, and 
$p \in \mathbb{R}^k$ the 
weights of all task-specific layers. 
%
Adapting $f$ by pruning amounts to finding 
the optimal binary mask $m^* \in \{0,1\}^d$,
such that 
applying $m^*$ to $w_0$ 
maximises the performance of $f$ on the 
target task.
%
%
The formal definition of the masking problem
is provided in Eq. \eqref{eq:main_loss} 
in \S\ref{sec:our_method}.
This is a high-dimensional combinatorial
optimisation problem, 
a highly challenging task in general.

\paragraph{Contributions}
Firstly, we propose an efficient 
adapting-by-pruning algorithm (\S\ref{sec:our_method}). 
We approximate mask searching 
as a differentiable optimisation problem,
prove that, under standard assumptions, the 
error of the approximation is bounded,
and propose an algorithm to efficiently 
find (near-)optimal masks. 
Our approach also allows the user to trade off between 
the performance and the efficiency of 
the pruned model by penalising the 
number of ones in $m^*$. 
The theoretical analysis and the proposed
algorithm are general
and, hence, define a generic model-adaptation paradigm.

Secondly, we test the effectiveness of our 
adapting-by-pruning algorithm by applying it
to BERT \cite{naacl19-bert} on 
multiple GLUE \cite{iclr19-glue} benchmark tasks.
%
Results (in \S\ref{sec:exp}) suggest that our method can
achieve the same performance 
as models adapted through feature-extraction but with (up to) 99.5\% fewer parameters.
Compared to fine-tuning, 
our method prunes up to 50\% parameters in BERT
while achieving comparable performance.
%
%
We also show that, 
with the same number of parameters,
sub-networks obtained by 
our method significantly
outperforms those obtained by other state-of-the-art 
pruning techniques.

Finally, we inspect the differences between the 
masks obtained by ours and existing
pruning methods, in particular the 
\emph{lottery ticket pruning methods} 
\cite{iclr19-lt,nips20-bert-lt}.
We find that the masks obtained by
our method and lottery-ticket approaches 
have very different topological structures,
but they also share some important characteristics,
e.g., their performance is highly sensitive to 
mask-shuffling and weight-reinitialisation \cite{arxiv-oneshot-bad}.
Furthermore, with ablation studies, we
find that \emph{connection recovery}, 
i.e., weights that are pruned in 
an early stage of the pruning process
are reactivated in a 
later stage, is a key ingredient for the
strong performance of our method,
especially at high-sparsity levels.
As connection recovery is not allowed in many 
popular pruning methods, 
this finding sheds light on the 
limitations of existing techniques.
Our codes and the Supplementary Material
can be found at \url{https://github.com/yg211/bert_nli}.

\section{Related Work}
\label{sec:related_work}
\paragraph{Model Compression and Pruning.}
Pruning is a widely used method to adapt pre-trained models
to downstream tasks while reducing the model size.
%
The \emph{lottery ticket hypothesis}, which lays the foundation
for many pruning works, states
that a sub-network can be fine-tuned in place of the
full model to reach comparable performance
\cite{iclr19-lt}. 
The optimal sub-network is also known as the 
\emph{winning ticket}.
Recent experiments have confirmed the existence 
of winning ticket sub-networks in different 
pre-trained models, including BERT \cite{icml20-bert-stablise}.

\emph{Iterative magnitude pruning} (IMP) \cite{iclr19-lt}
is the most widely used 
method for finding winning tickets.
IMP involves three steps at each iteration:
(i) training the network until convergence,
(ii) removing a small percentage of connections 
(usually 10-20\%)
whose weights are closest to zero, and
(iii) rewinding all remaining weights 
to their initial values. 
Such iterations are repeated until the target
sparsity is reached.
IMP is highly expensive 
because each iteration requires training the model to
convergence and the obtained pruned network
needs to be further fine-tuned.
Cheaper alternative methods have been proposed 
\cite{nips20-bert-lt,acl20-bert-translate,nips20-continuous-sparsification}, 
but the quality of their sub-networks 
are worse than IMP, and the obtained
sub-networks also need to be fine-tuned.
Compared to these pruning techniques, our method keeps all original weights intact, hence
avoids the expensive fine-tuning step
and allows for parameter reuse.

Other model compressing methods include 
weight quantization \cite{aaai20-qbert,iclr20-quantization}, 
parameter sharing 
\cite{iclr20-albert} and attention layer decomposition \cite{acl20-deformer}.
Because our method reuses the original neural structure,
these methods can be 
used together with our paradigm 
to further reduce the model size and 
the inference time.

\paragraph{Supermask \& Neural Architecture Search.}
Our work is inspired by works on 
\emph{supermasks} such as \cite{nips19-supermask}.
They find that applying appropriate binary masks
to randomly initialised neural networks can improve the performance of the original model, 
even without updating the weights values.
These binary masks are known as supermasks.
\cite{nips19-supermask} formulates the supermask learning problem
as the problem of learning a Bernoulli distribution for each weight;
at training time, they learn the distribution 
with standard stochastic gradient descent
(SGD); in forward propagation,
the activation status of each weight is sampled from the learnt distribution stochastically.
However, in this case, users cannot explicitly control the sparsity
of the pruned network.

\citet{cvpr20-binary} proposed an improved 
algorithm, \emph{edge-popup}, 
which removes the stochasticity by 
ranking the importance of neural connections.
Edge-popup provides better performance
with smaller variance and allows 
users to specify the target sparsity. 
However, edge-popup was only tested on models
with randomly initialised weights, and no proofs were
provided regarding its optimality.
Instead of learning a score to rank
the neural connections, our method directly learns
the $\{0,1\}$ values in the mask; this allows us to prove 
the optimality guarantees. 
Also, we apply our method to prune a pre-trained model
(BERT) instead of the random-initialised models, 
and provide the first systematic comparison
of the topological structures of supermasks and 
winning lottery tickets (in \S\ref{subsec:topology_analysis}).

\section{Our Method}
\label{sec:our_method}
Let $f(x;w_0,p)$ denote the model
for a downstream task, where $w_0$ are the
weights from the pre-trained base model 
and $p$ the task-specific parameters. 
%
Our target is to find 
a task-specific optimal binary mask (\emph{supermask}) for $w_0$.
More formally, a supermask $m^*$ can be defined as
\begin{align}
\label{eq:main_loss}
&m^* = {\arg\min}_{m \in \{0,1\}^d} 
\min_{p \in \mathbb{R}^k} \mathcal{L}(m,p),
\nonumber
\\
&\mathcal{L}(m,p) = |D_{train}|^{-1} 
\sum_{z \in D_{train}}
\ell(z; m, p), \\
&\ell(z; m, p) = 
\mathcal{G}\left(f(x; w_0 \!\odot\! m, p), y\right),
z = (x, y). \nonumber
\end{align}
where $\odot$ is the element-wise product and
$\mathcal{G}(\hat{y},y)$ a differentiable loss function, 
e.g., the least-squares or cross-entropy loss. 
Because $\mathcal{L}$ is 
differentiable with respect to its arguments,
$p$ can be optimised with standard SGD techniques.
However, as $m \in \{0,1\}^d$ is a discrete variable, 
we cannot directly use SGD approaches to 
optimise $\mathcal{L}$ with respect to $m$.
In \S\ref{subsec:appr},
we propose a continuous approximation
of $\mathcal{L}$ 
and  in \S\ref{subsec:proof} we 
prove that (near-)optimal supermasks can be 
obtained by optimising such a continuous 
approximation;
in \S\ref{subsec:algorithm},
we propose an algorithm to obtain supermasks
of arbitrary sparsity.

\subsection{Continuous Approximation}
\label{subsec:appr}
A natural method to approximate binary
masks $m \in \{0,1\}^d$ is to use the 
sigmoid function, i.e., to replace $m$ with 
$$m_{t} = \sigma(t\theta) = [1+\exp(-t \theta)]^{-1}, \quad t>>1$$ 
where $\theta \in \mathbb{R}^d$ 
is a real-valued tensor
of the same dimension as $m$, and 
$t$ is a hyper-parameter.
With this approximation, the loss in Eq. \eqref{eq:main_loss} becomes
a continuous function, i.e., 
\begin{align}
\label{eq:appr_loss}
{\mathcal{L}}_t(\theta) 
= |D_{train}|^{-1} 
\!\!\!
\sum_{z\in D_{train}}
\!\!\!
\ell(z; m_t, p).
\end{align}
Since $m_t$ is differentiable with respect to 
$\theta$, the gradient of \eqref{eq:appr_loss} is well 
defined as
\begin{align}
{\nabla}_{\theta} & {\mathcal{L}}_{t}(\theta)
   = \nonumber \\ 
& |D_{train}|^{-1} 
\!\!\!\!\!
\sum_{z\in D_{train}}
\!\!\!\!\!
\nabla_{m_t} \ell(z;m_{t},p) \odot 
\nabla_\theta m_{t}.
\label{eq:single_t}
\end{align}
When $t \to \infty$,  we have
$m_t \to m$  and ${\mathcal L}_{t} \to {\mathcal L}$, 
i.e., the gap between the original objective 
function and its continuous approximation approaches zero.
However,  for any $\theta \neq 0$, $t \to\infty$  also implies 
$\nabla_{\theta} m_t = m_t \odot (1 - m_t)\to 0$ 
and hence the gradient in Eq. \eqref{eq:single_t}
also vanishes.
To trade off between the quality of the 
approximation (Eq. \eqref{eq:appr_loss}) 
and the magnitude of its gradient (Eq. \eqref{eq:single_t}), 
we propose to\footnote{The 
technique can be seen as a 
variation of the \emph{BinaryConnect} 
approach described in 
\cite{nips15-binnary-connect}. Convergence guarantees for the 
BinaryConnect method can be found in 
\cite{li2017training}, but their stochastic 
approach produces an irreducible convergence gap 
that our method can avoid.
}
use a larger $t$ in  
the forward propagation and a smaller $t$
in sigmoid gradient computation. 
More explicitly, we approximate Eq. \eqref{eq:single_t} with
\begin{align}
\label{eq:double_t}
&\Tilde{\nabla}_{\theta} {\mathcal{L}}_{t}(\theta)
=  |D_{train}|^{-1} 
\!\!\!\!\!
\sum_{z\in D_{train}}
\!\!\!\!\!
\Tilde{\nabla}_{\theta} \ell(z;m_{t_l}, m_{t_s},p) 
\nonumber
\\
&\Tilde \nabla_{\theta} \ell(z;m_{t_l}, m_{t_s},p) = 
\nabla_{m_{t_l}}\ell(z;m_{t_l},p) \odot 
\nabla_{\theta} m_{t_s},
\end{align}
where $t_l$ and $t_s$ stand for the 
large and small values of $t$, respectively.
The approximate SGD updates of $\theta$ and $p$ are 
\begin{align}
\label{eq:sgd_theta}
\theta_{i+1} &:= \theta_{i} - \alpha_i 
\tilde \nabla_{\theta}\ell(z; m_{t_l}, m_{t_s}, p),
\\ 
\label{eq:sgd_p}
p_{i+1} &:= p_i - \beta_i \nabla_p \ell(z; m, p),
\end{align}
where $\alpha_i$ and $\beta_i$ are learning steps, 
$\Tilde \nabla_{\theta} \ell$ is defined in Eq. (\ref{eq:double_t}),  
$\nabla_p \ell$ is the exact gradient of $\ell$
(in Eq. \eqref{eq:main_loss})
with respect to $p$, 
and $z$ is chosen uniformly at random in $D_{train} $.
After $T$ steps, a binary mask can
be derived from $\theta_T$ by forcing
all positive values in $\theta_T$ to be
one and all negative values to be zero
i.e., by letting 
$m_T \to \mathbbm{1}[\theta_T > 0]$,
where $\mathbbm{1}$ is the indicator function.

\subsection{Convergence analysis}
\label{subsec:proof}
Let $v \in [0,1]^d$ denote an approximate mask, i.e.,
$v = \sigma(t \theta)$ for some $t > 0$ and 
$\theta \in {\mathbb R}^d$.
We assume that $\ell$ is convex 
in $v$ for all $z$, and that $\nabla_v \ell$ 
is bounded, i.e., that 
$\ell$ satisfies the assumption below.
\begin{assumption}
	\label{assumptions ell}
	For all $v, v' \in [0, 1]^d$, any possible 
	input $z$ and any parameters $p \in \mathbb{R}^k$, 
	$\ell(v, z)$ is differentiable with respect to $v$ and 
	\begin{align*}
		&\max_{z,v, p} 
		\| \nabla_v \ell(z;v,p) \|^2 \leq G^2, \\
		&\ell(z;v,p) - \ell(z;v',p) \geq \nabla_{v'}
		\ell(z;v',p)^\intercal
		(v - v'), 
    \end{align*}
	 where $G$ is a positive constant.
\end{assumption}

\noindent 
Under this assumption\footnote{
Assuming a certain regularity of the objective 
is standard practice for proving SGD 
convergence bounds. 
}, 
we prove that the SGD steps in 
Eq. \eqref{eq:sgd_theta} 
produce a near-optimal mask.

\begin{theorem}
\label{theorem:main}
Let $\ell$ meet Assumption \ref{assumptions ell} and
$\{ \theta_i \in {\mathbb R}^d\}_{i=1}^T$ be the sequences of stochastic weight 
updates defined in Eq. \eqref{eq:sgd_theta}. 
For $i=1, \dots, T$, let 
$z_i$ be a sample drawn from $D_{train}$ uniformly 
at random, and
$\alpha_i = \frac{c}{\sqrt{i}}$, where $c$ is a 
positive constant. Then 
\begin{align*}
\mathbb{E}[{\cal L}_{t}(\theta_T) - &
{\cal L}_{t}(\theta^*)] 
\leq \\ 
& \frac{1}{c \sqrt{T}} + 
\frac{c G^2 (1 + C) (1 + \log T)}{T},
\end{align*}
where the expectation is over the 
distribution generating the data,
$\theta^* = \arg \min_{\theta \in \mathbb{R}^d} 
{\cal L}_{t_l}(\theta)$, 
$G$ is defined in Assumption \ref{assumptions ell} 
(with $v = \sigma(t_l \theta)$), 
and 
	\begin{align*}
		&C = t_l t_s \left(\frac{1}{t_l t_s}  
	 - 2 g_{max}(t_l) g_{max}(t_s)
	  + \frac{t_l t_s}{16^2} \right)
	\\
		&g_{max}(t)  
    = \sigma(t M)
    (1 - \sigma(t M)), \nonumber 
	\end{align*}
	with $M = \max \{ |\theta_i|, i=1, \dots, T \}$.
\end{theorem}

Proof of Theorem~\ref{theorem:main} is provided
in the Supplementary Material.
Theorem \ref{theorem:main} suggests that the expected gap
between the loss evaluated at $\theta_T$
and the true minimum $\theta^*$ is bounded.
%
For simplicity, we consider 
the convergence of Eq. \eqref{eq:sgd_theta} for 
fixed $p$, but a full convergence proof 
can be obtained by combining 
Theorem \ref{theorem:main} with standard results 
for unconstrained SGD.
By combining the Theorem \ref{theorem:main} with 
$\lim_{t_l \to \infty}{\mathcal L}_{t_l} = {\mathcal L}$, 
we conclude that a near-optimal 
supermask can be defined by  
$$m^* = \lim_{t_l \to \infty} \sigma(t_l\theta_T) 
= \mathbbm{1}[\theta_T > 0].$$

\subsection{Adapt by Pruning Algorithm}
\label{subsec:algorithm}
We can use SGD with Eq. 
\eqref{eq:sgd_theta} and \eqref{eq:sgd_p} 
to obtain a
near-optimal binary mask of undefined sparsity.
Note that the number of zeros in the
approximated mask $\sigma(t\theta_T)$
is determined by the number of negative values
in $\theta_T$.
To encourage more negative values in $\theta$,
we add a regularisation term in 
Eq. \eqref{eq:sgd_theta}: %
\begin{align}
\theta_{i+1} := \theta_{i} - \alpha_i 
\Tilde{\nabla}_\theta \ell(z;m_{t_l}, m_{t_s},p) - 
\gamma_i \mathbf{1},
\label{eq:sgd_sparsity}
\end{align}
where $\mathbf{1}$
is a tensor of the same dimension as $\theta$
and its elements are all $1$s,
and $\gamma_i$ is the weight for the regularisation 
term\footnote{This is equivalent to adding 
$\gamma 1^\intercal \theta$ to ${\cal L}_t(\theta)$
in Eq. \eqref{eq:appr_loss}.}.
In practice, we can initialise  $\theta$
with small positive values;
the penalisation term will push the entries 
of $\theta$ to be negative, so that  
sparser masks can be obtained. 
To ensure that the target sparsity can be
achieved with a similar number of epochs
as regular training (i.e., one-shot pruning),
the value of the hyper-parameter
$\gamma_i$ should be tuned to
control the pruning speed. 
Alg. \ref{alg:alg} presents the pseudo
code for the our supermask searching algorithm. 

\begin{algorithm}[H]
\caption{Adapt by Pruning Algorithm
\label{alg:alg}
}
\textbf{Inputs:} 
Pre-trained model $f(x;w_0,\cdot)$,
training data $D_{train}$, target sparsity $s$,
hyper-parameters $t_l$ and $t_s$,
initial parameters $\theta_0$ and $p_0$, learning steps
$\alpha_i, \beta_i$ and $\gamma_i$
for $i=1, \cdots, T$ \\
\\
\textbf{Pruning:}
\begin{algorithmic}
\FOR{$i = 1, \cdots, T$}
    \STATE Obtain $\theta_i$ following Eq. \eqref{eq:sgd_sparsity}
    \STATE Obtain $p_i$ following Eq. \eqref{eq:sgd_p}
    \STATE \textbf{break} if ratio of
    negative values in $\theta_i$ is larger than $s$
\ENDFOR
\end{algorithmic}
\textbf{\\Output:}
$p_i$ and the binary mask $m^* = \mathbbm{1}[\theta_i > 0]$
\end{algorithm}

\section{Experiments}
\label{sec:exp}

In this section, we test the efficiency of our pruning method by applying it to
BERT-base.
The BERT model and its weights are from the
PyTorch implementation by HuggingFace\footnote{\url{https://github.com/huggingface/transformers}}.
All our experiments are performed on a workstation
with a single RTX2080 GPU card (8GB memory).

We consider three representative tasks
from the GLUE benchmark \cite{iclr19-glue}:
STS-B (sentence similarity measurement),
SST-2 (sentiment analysis),
and MNLI (natural language inference).
They cover a wide range of input types
(single or pairwise sentences), 
output types (regression or classification),
and data size; 
see Table \ref{tab:datasets} for their details.
Since the official test set are not publicly available,
we report the performance on their official Dev set.

\begin{table}[h]
    \small
    \centering
    \begin{tabular}{p{0.65cm} p{2cm} p{1.6cm} p{1.6cm} }
    \toprule
    & STS-B & SST-2 & MNLI \\
    \midrule
    Train & 7k & 67k & 393k\\
    Dev & 1.5k & 872 & 20k \\
    Input & SentPair & SingleSent & Sentpair \\
    Output & Score $\in [0,5]$ & 2-class label & 3-class label\\
    Metric & Spearman's rho & Accuracy & Accuracy\\
    \bottomrule  
    \end{tabular}
    \caption{GLUE datasets used in the experiments. 
    }
    \label{tab:datasets}
\end{table}

For each task, we learn a weighted average of each layer's
[CLS] vector as the input to the task-specific MLP.
The MLP has a ReLU hidden layer of the same dimension
of the input (i.e., 768), followed by the output layer.
\cite{rep4nlp19-adapt} have shown that
using the weighted average of all layers outputs 
yields better performance than only using 
the last layer's [CLS].
Parameters in the added MLP account for around 0.5\%
of the whole model size, and they are not pruned
in our experiments, in line with other BERT pruning methods
\cite{nips20-bert-lt,acl20-bert-translate}.

\paragraph{Baselines}
We compare our method with both feature extraction and
fine-tuning, the two predominant paradigms of
model adaptation. The performance of fine-tuning is
widely regarded as the `upper-bound' in model adaptation
\cite{icml19-adaptor} and model compression \cite{arxiv20-compress-bert-survey}.
We also consider three pruning methods as baselines.
\textbf{Rnd} is the random baseline:
for a given sparsity level $s$, 
this method builds a mask such that each element 
is 0 with probability $s$ and 1 otherwise.
\textbf{IMP} is the iterative magnitude pruning method
(see \S\ref{sec:related_work});
it is the most widely used heuristic
for obtaining winning tickets, but is also highly
expensive as it only prunes 20\% remaining weights
in each iteration.
\textbf{MP} is the magnitude pruning method
proposed by \cite{iclr18-mag-prune};
unlike IMP that waits until convergence to prune
weights, MP prunes weights every few hundred 
SGD steps. It is widely regarded
as the best one-shot pruning method 
\cite{arxiv-mp-best,arxiv-oneshot-bad}.
Note that our algorithm also falls into the
one-shot pruning category as it avoids
iterative pruning.

\paragraph{Hyper-parameters}
For feature extraction and fine-tuning, we
use the same setup as in \cite{rep4nlp19-adapt},
including their learning rate scheduling scheme
and all hyper-parameters.
For the pruning baselines, we use the same
setup as in \cite{nips20-bert-lt}.
%
For our algorithm (Alg. \ref{alg:alg}),
we randomly sample 10\% data from the training set
of each task as the validation set, and
we perform hyper-parameter grid-search to select the
ones with best performance on the validation set.
%
%
To initialise $\theta_0$ (Eq. \eqref{eq:appr_loss}),
the entries of $\theta_0$ are 
drawn from a normal distribution
with mean $0.01$ and variance $0.001$;
other initialisation strategies that 
assign only positive
values (e.g., to use a uniform distribution, 
or assign the same value to all 
elements) yield similar performance.
When values in $\theta_0$ are negative,
the corresponding weights in BERT 
are `pruned' at the beginning, which 
harms the performance of the final supermask.
More details about the hyper-parameter choices 
and the selection process are presented 
in the Supplementary Material.

\subsection{Main Results}
\label{subsec:wo_finetuning}

\begin{figure*}
    \centering
    \begin{subfigure}[b]{0.31\textwidth}
         \centering
         \includegraphics[width=\textwidth]{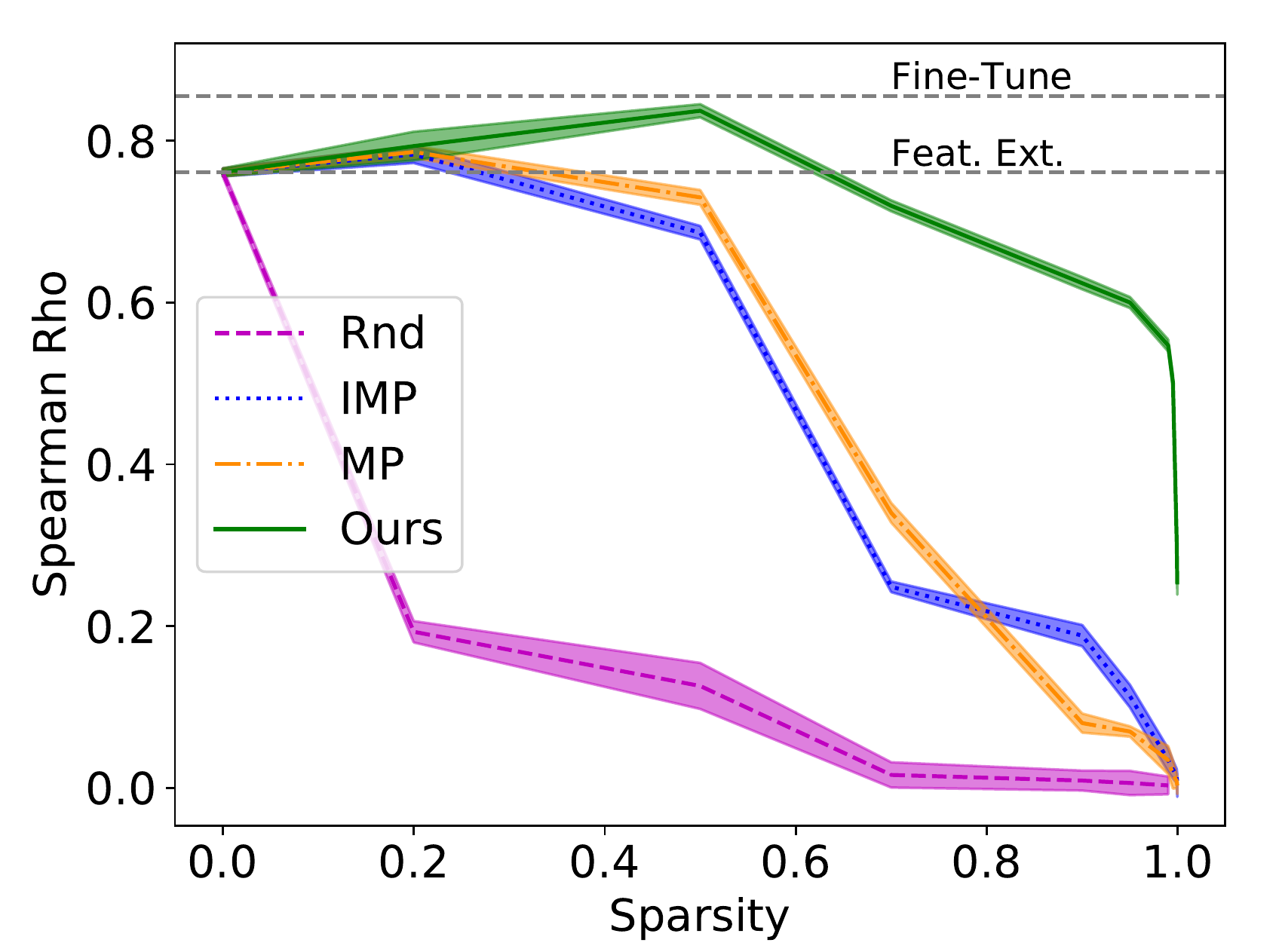}
         \caption{STS-B}
     \end{subfigure}
     \begin{subfigure}[b]{0.31\textwidth}
         \centering
         \includegraphics[width=\textwidth]{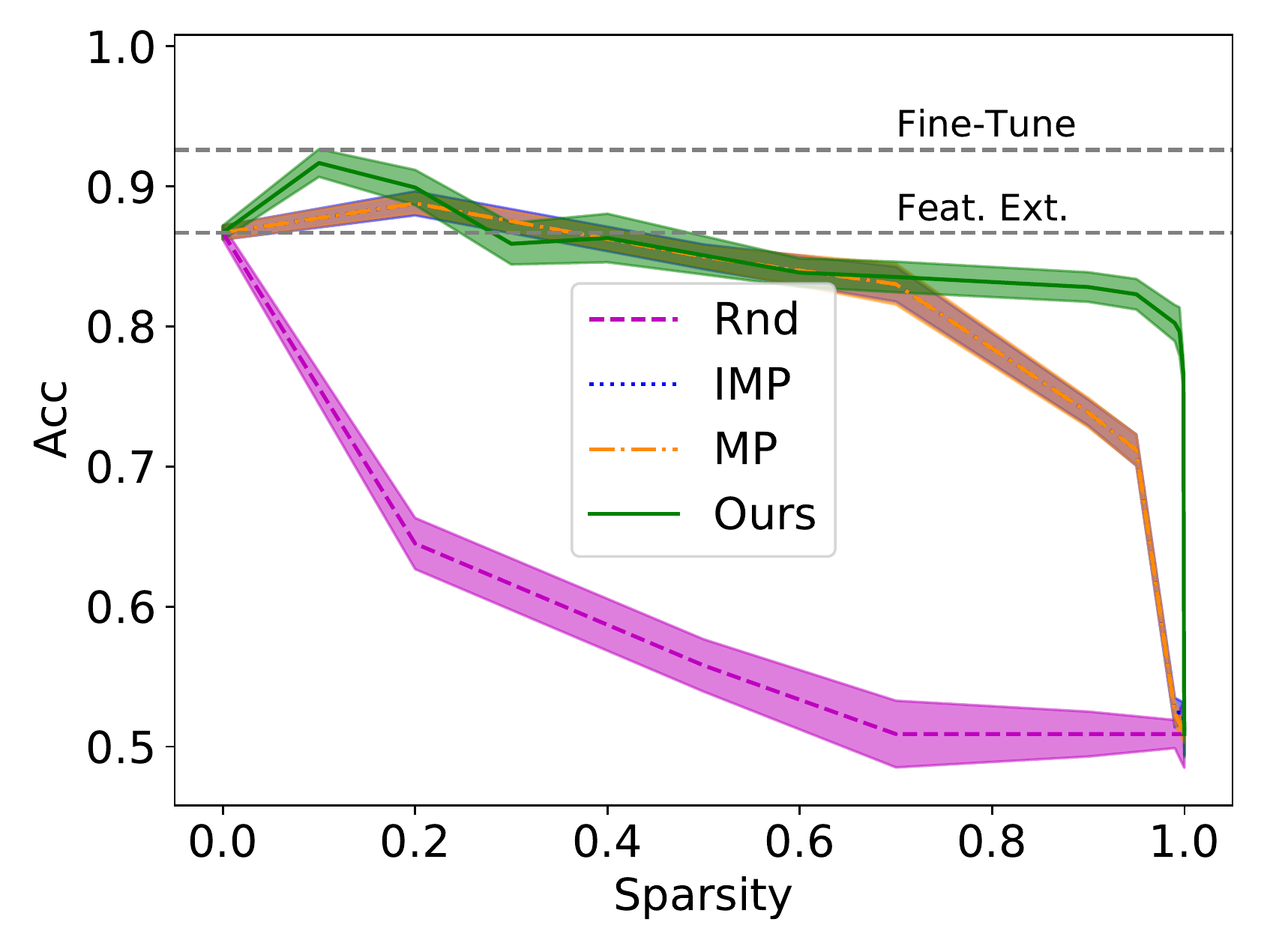}
         \caption{SST-2}
     \end{subfigure}
     \begin{subfigure}[b]{0.31\textwidth}
         \centering
         \includegraphics[width=\textwidth]{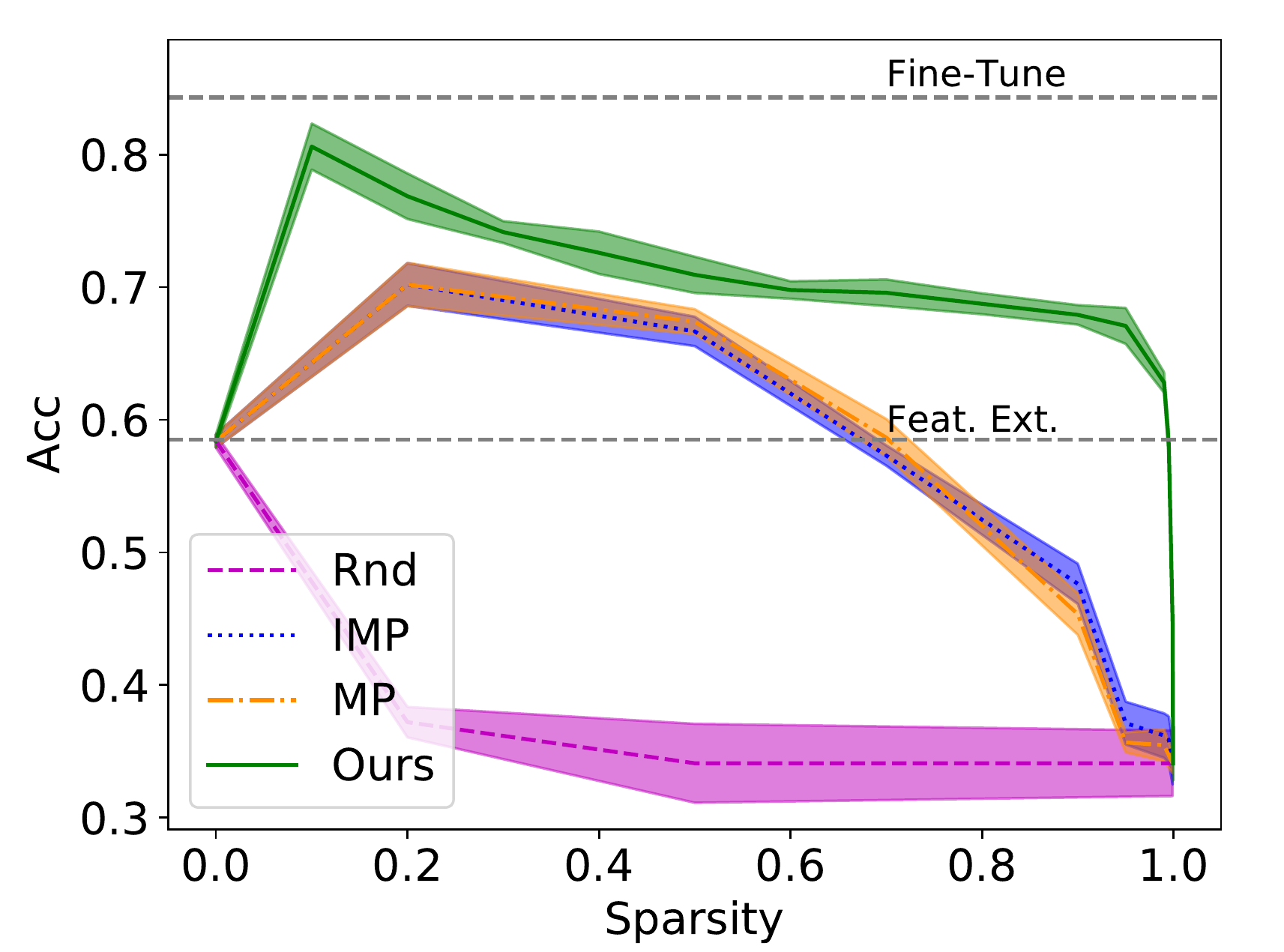}
         \caption{MNLI}
     \end{subfigure}
     \caption{
     Comparison of multiple model adaptation and
     pruning methods.
     %
     All results are averaged over five runs with different random seeds; shaded areas represent 
     the corresponding standard deviations.
     Detailed performance can be found in the Supplementary
     Material.}
     \label{fig:main_results}
\end{figure*}

Fig. \ref{fig:main_results} compares the performance
of our approach and other model adaptation 
and pruning methods.
Compared to feature-extraction,
in all three tasks, our method can obtain 
sparse networks with significantly\footnote{We
test significance with
double-tailed t-test with $p < 0.01$.} better performance;
particularly, in MNLI, our method can prune up to
99.5\% parameters of BERT while reaching 
better or comparable performance. 
Compared to fine-tuning, our method can
extract sub-networks with comparable performance,
without statistically significant gaps in STS-B and SST-2.
These observations suggest that our method
shares the advantages of feature extraction and
fine-tuning (re-use weights and structures),
but can also reduce the model size (see Table \ref{table:best_of_both_world}).

As for the pruning strategies, 
to ensure fair comparison,
we keep all un-pruned weights from BERT with
their original values 
(see the bottom row in Fig. \ref{fig:motivation}).
We find that the performance of all 
pruning methods drops as the sparsity level
increases, but our method outputs
the best-performing sub-networks across 
all sparsity levels and tasks, and the performance gap
is significant under most conditions.
Furthermore, the performance of our method
is more robust to the growth of sparsity:
the other pruning methods encounter a rapid
performance drop when the sparsity reaches
0.5 -- 0.7, but our method mostly remains its performance until the sparsity reaches 0.95. 
These observations show that our approach can 
consistently produce (near-)optimal supermasks for BERT
(see \S\ref{sec:our_method}).

\subsection{Topological Analysis}
\label{subsec:topology_analysis}
\begin{figure}
    \vspace*{-12pt}
    \centering
    \begin{subfigure}[b]{0.5\textwidth}
         \centering
         \includegraphics[width=0.48\textwidth]{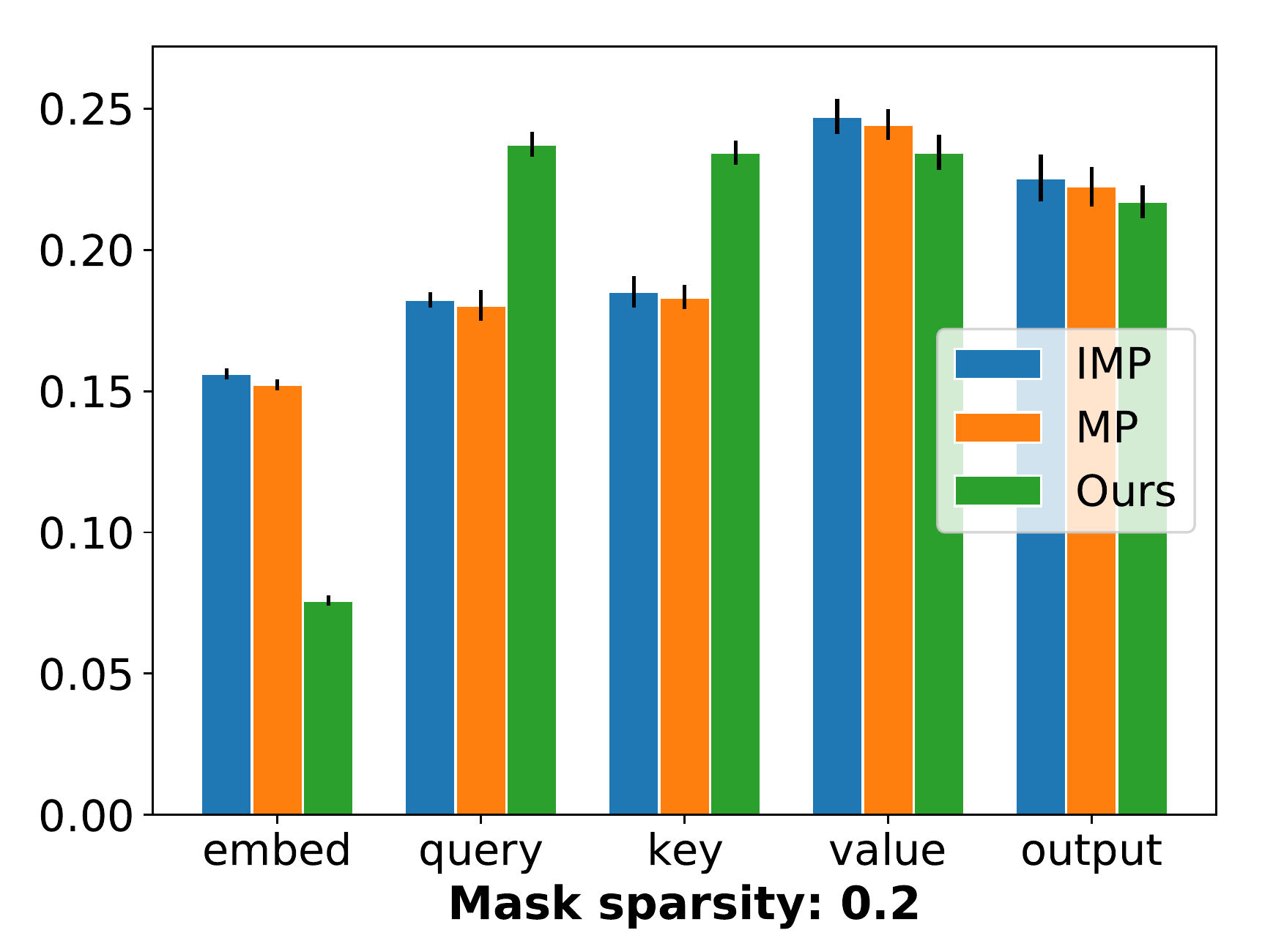}
         \includegraphics[width=0.48\textwidth]{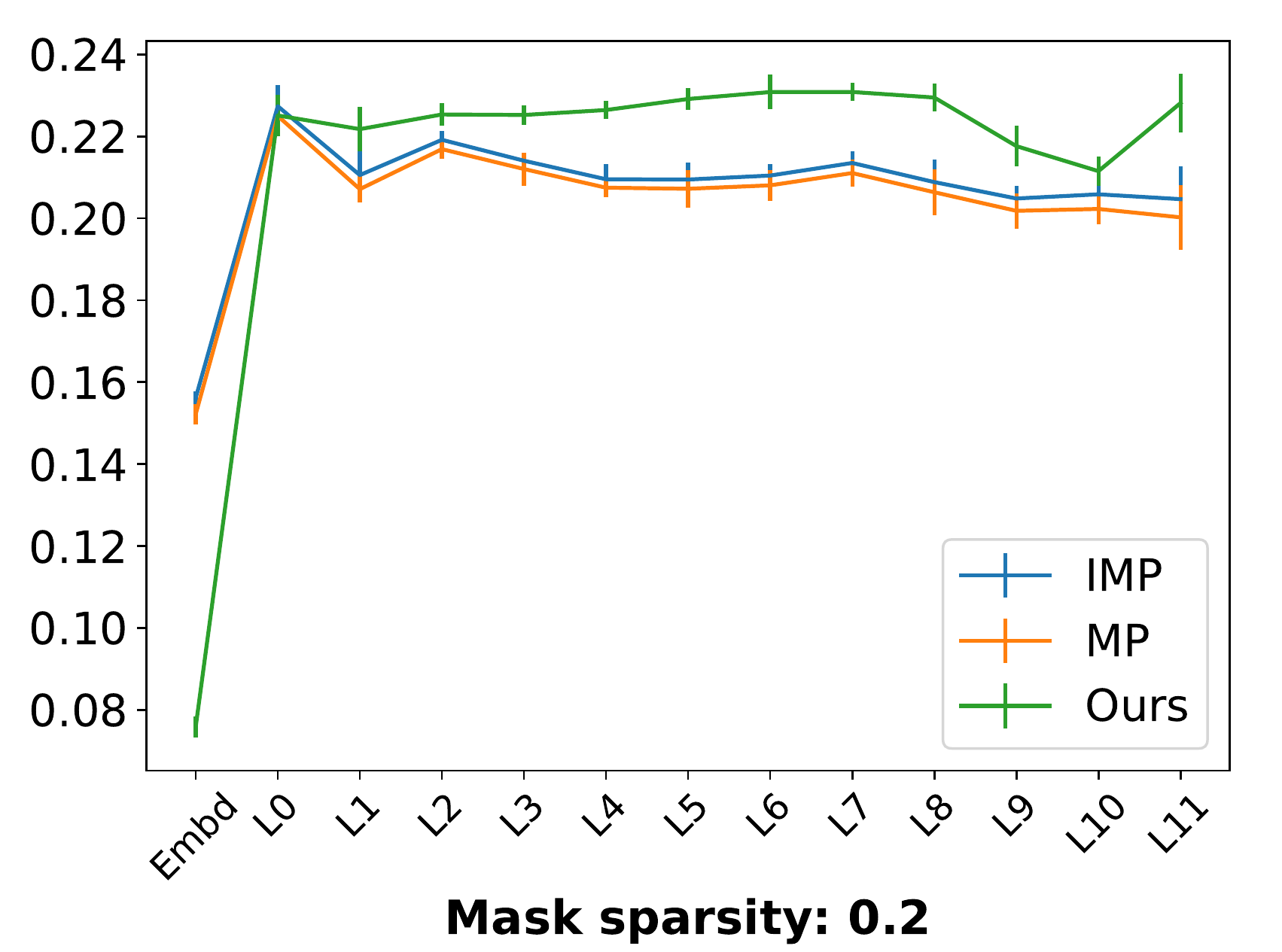}
     \end{subfigure}
     %
     %
     \begin{subfigure}[b]{0.5\textwidth}
         \centering
         \includegraphics[width=0.48\textwidth]{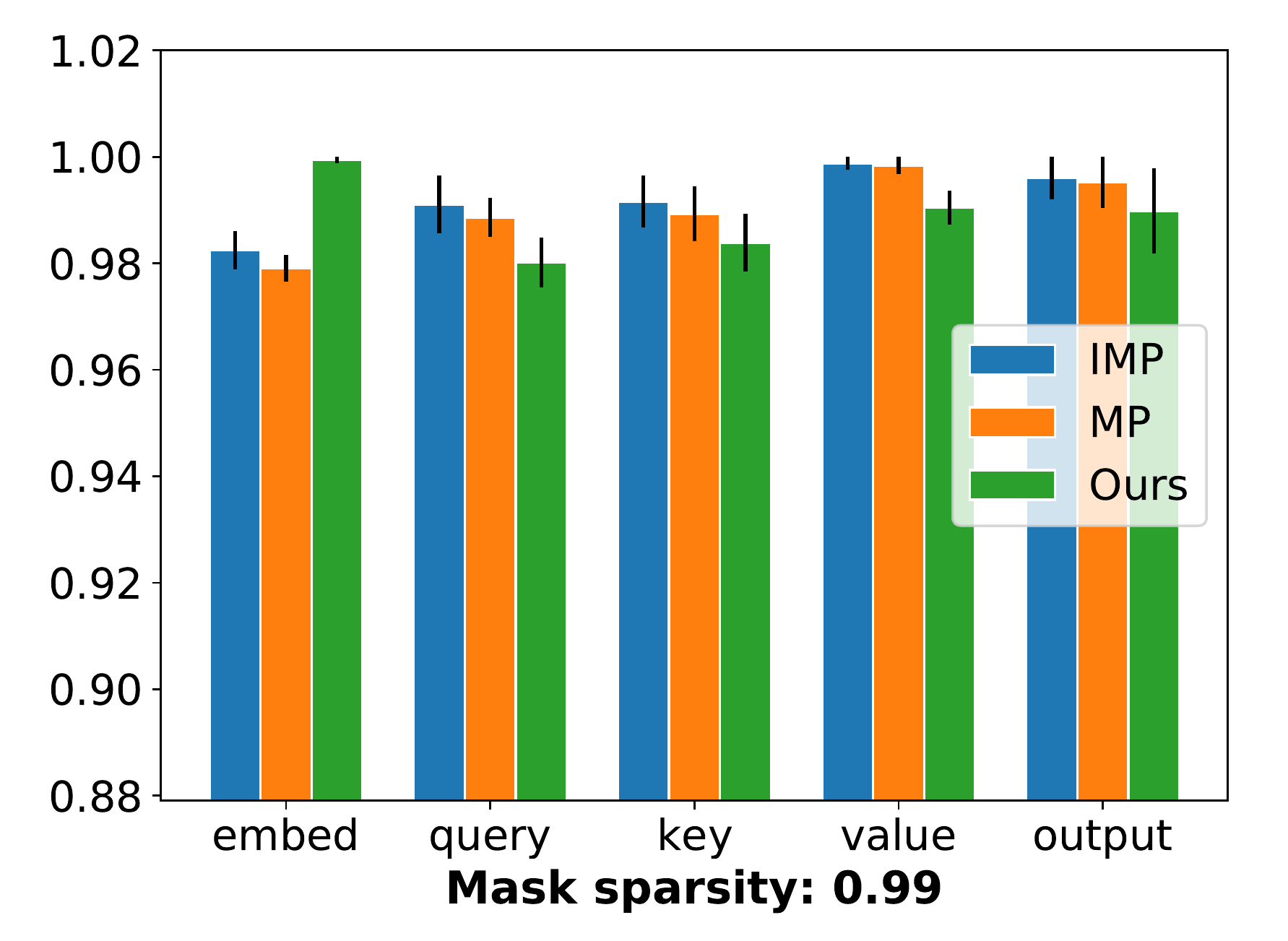}
         \includegraphics[width=0.48\textwidth]{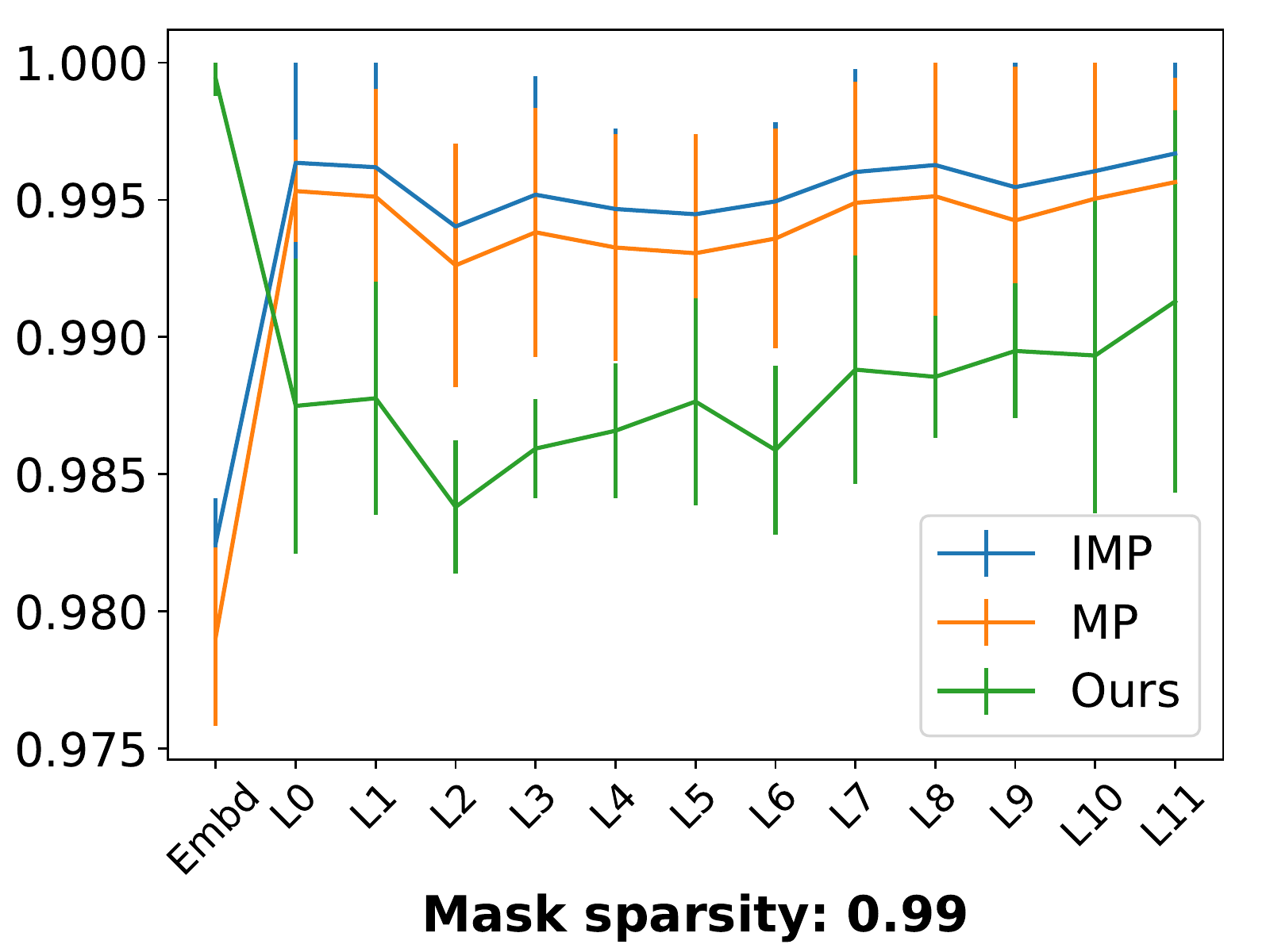}
     \end{subfigure}
     \caption{
     The distribution of the 
     surviving connections over 
     different components (left) and 
     layers (right) is different 
     in the sub-networks obtained through our approach
     or MP/IMP.
    %
    Results obtained from five runs on MNLI. 
     Results for the other datasets are
     in Supplementary Material.
    Error bars: standard deviation.
     }
     \label{fig:sparsity_distribution}
\end{figure}

In this subsection, we inspect
the topological differences between supermasks 
(obtained by our method) and task-specific winning tickets
(obtained by IMP/IMP).
Fig. \ref{fig:sparsity_distribution} 
shows that that, even at the same sparsity levels,
supermasks are substantially different
from winning tickets, in terms of layer-wise 
sparsity and component-wise sparsity.
For example, at sparsity 0.2 
(top row in Fig. \ref{fig:sparsity_distribution}),
IMP/MP prunes more weights in the embedding layer
and fewer in the Transformer layers and components;
but at higher sparsity levels 
(0.99, bottom row in 
Fig. \ref{fig:sparsity_distribution}),
the pattern is reversed: 
compared to IMP/MP, our method prunes
more weights in the embedding layers and 
fewer in Transformer layers.
Similar observations can be made
consistently across all selected GLUE tasks.
These results suggest that
(i) the topological structures of supermasks and 
winning tickets are very different, and 
(ii) the topological differences  
should follow from differences 
in the pruning strategy
rather than differences in the downstream tasks.
The causality and relation between 
layer/component-wise sparsity and the model
performance is worth further investigation,
and we leave it for future work.

\begin{table}[]
    \centering
    \small
    \begin{tabular}{l | r r | r r | r r}
    \toprule
    Spa-
    & \multicolumn{2}{c|}{STS-B (rho)}   
    & \multicolumn{2}{c|}{SST-2 (acc)}   
    & \multicolumn{2}{c}{MNLI (acc)} \\
    rsity 
    & Shufl & Reinit & Shufl & Reinit 
    & Shufl & Reinit \\
    \midrule
    0.2 & .397 & 0.05 & .826 & .491 & .714 & .328 \\
    0.5 & .312 & -.019 & .808 & .491 & .601 & .328 \\
    0.7 & .282 & .006 & .787 & .491 & .554 & .328 \\
    0.9 & .268 & -.036 & .720 & .491 & .403 & .328 \\
    0.95 & .213 & .008 & .589 & .491 & .318 & .328 \\
    0.99 & .155 & -.016 & .509 & .491 & .353 & .328 \\
    0.999 & .011 & .005 & .509 & .491 & .343 & .328 \\
    \bottomrule
    \end{tabular}
    \caption{Supermasks are sensitive to shuffling and 
    weight re-initialisation.
    For re-initialisation, each pre-trained
    weight is multiplied with a random value drawn from 
    normal distribution $\mathcal{N}(1,\sigma)$,
    where $\sigma = 0.01$. Performance with 
    $\sigma = 0.1$ and $0.001$ are similar.
    }
    \label{tab:shuffle_reinit}
\end{table}

\subsection{Ablation Study}
\label{subsec:ablation_study}
To further understand what contributes to the
strong performance of supermasks obtained by
our method, we perform multiple ablation study
in this subsection.

\paragraph{Sensitivity Test}
Winning tickets obtained by IMP and MP
are sensitive to 
\emph{mask shuffling} (where 
the surviving connections of a given 
layer are randomly redistributed over that layer)
and 
\emph{weight re-initialisation} 
(where $w_0$ is rescaled before fine-tuning) \cite{iclr19-lt,arxiv-oneshot-bad}.
We test whether supermasks share these 
characteristics
by performing shuffling and re-initialisation operations
on the obtained supermasks, applying the new masks
to BERT and re-update the weights in the
task-specific MLP.
From Table \ref{tab:shuffle_reinit},
it is clear that supermasks are sensitive to 
shuffling and, especially, weight re-initialisation;
even small disturbance on the weights 
drops the performance to near-random level.
This finding suggests that our method 
is able to pick the
most important weights in the original model
(c.f., \cite{arxiv-oneshot-bad}).

\paragraph{Connection Recovery}
In IMP/MP, if a weight
is pruned in an early stag, it will remain pruned
in all follow-up steps. In our algorithm,
negative elements of $\theta$ can become
positive at later stages. 
Hence, our method allows for pruned  
network connections to be
reactivated while IMP/MP does not.

A considerable percentage of weights are
reactivated in our algorithm:
for example, in a sparsity-0.99 sub-network
obtained by our method on MNLI,
12.3\% of its connections were recovered from 
the connections pruned in early stages.
To study how weight reactivation
affects the performance of supermasks, 
we revise Alg. \ref{alg:alg} 
so that elements that become negative 
are forced to remain negative
in all following SGD steps.
Performance of the revised algorithm is presented
in Table \ref{tab:mask-nr}. Disabling connection
recovery significantly harms the performance 
across all tasks and sparsity levels.
With the growth of sparsity, the performance loss
also increases, suggesting that connection recovery
is particularly important for obtaining 
good high-sparsity models.
We believe this also explains the observation from
Fig. \ref{fig:main_results} that the gap between 
our method and IMP/MP increases with the growth
of sparsity. 

\begin{table}[]
    \vspace*{-9pt}
    \centering
    \small
    \begin{tabular}{l | r r | r r | r r}
    \toprule
    & \multicolumn{2}{c|}{STS-B }   
    & \multicolumn{2}{c|}{SST-2 }   
    & \multicolumn{2}{c}{MNLI} \\
    Sparsity 
    & Rho & $\Delta$ 
    & Acc  & $\Delta$ 
    & Acc & $\Delta$ \\
    \midrule
    0.2 & .768 & -.026 & .860 & .-.048 & .742 & -.027  \\
    0.5 & .804 & -.033 & .815 & .-.041 & .690 & -.019 \\
    0.7 & .628 & -.092 & .803 & -.040 & .653 & -.043 \\
    0.9 & .563 & -.061 & .790 & -.046 & .628 & -.051 \\
    0.95 & .519 & -.081 & .783 & -.048 & .621 & -.050 \\
    0.99 & .417 & -.013 & .759 & -.051 & .575 & -.053 \\
    0.999 & .066 & -. 235 & .671 & -.101 & .333 & -.113 \\
    \bottomrule
    \end{tabular}
    \caption{Disabling connection recovery
    harms the performance. $\Delta$: 
    performance difference caused by disabling
    recovery, compared to sub-networks obtained by the standard Alg.
    \ref{alg:alg} at the same sparsity.}
    \label{tab:mask-nr}
\end{table}

\section{Conclusion}
\label{sec:conclusion}
In this work, we proposed a novel paradigm to adapt
pre-trained models to downstream tasks,
which prunes task-irrelevant connections while
keeping all remaining weights intact.
We formulated the pruning problem as an optimisation
problem and proposed an efficient algorithm
to obtain the (near-)optimal sub-networks.
%
%
%
Experiments on BERT showed that 
our method
can achieve highly competitive performance 
on downstream tasks, even without fine-tuning.
As it reuses the original architecture 
and weight values,
our method is particularly suitable for 
adapting a base model to downstream tasks with minimum extra data to download.
Moreover, pruning all task-irrelevant connections may help understand what are the `core' neural 
connections of the pre-trained models 
and provide inspirations to design new neural
architectures.
%

\bibliographystyle{acl_natbib}
\bibliography{emnlp2021} 

\begin{thebibliography}{28}
\expandafter\ifx\csname natexlab\endcsname\relax\def\natexlab#1{#1}\fi

\bibitem[{Blalock et~al.(2020)Blalock, Ortiz, Frankle, and
  Guttag}]{arxiv-prune-survey}
Davis Blalock, Jose Javier~Gonzalez Ortiz, Jonathan Frankle, and John Guttag.
  2020.
\newblock \href {http://arxiv.org/abs/2003.03033} {What is the state of neural
  network pruning?}
\newblock \emph{arXiv:2003.03033}.

\bibitem[{Brix et~al.(2020)Brix, Bahar, and Ney}]{acl20-bert-translate}
Christopher Brix, Parnia Bahar, and Hermann Ney. 2020.
\newblock \href {https://www.aclweb.org/anthology/2020.acl-main.360/}
  {Successfully applying the stabilized lottery ticket hypothesis to the
  transformer architecture}.
\newblock In \emph{ACL}.

\bibitem[{Cao et~al.(2020)Cao, Trivedi, Balasubramanian, and
  Balasubramanian}]{acl20-deformer}
Qingqing Cao, Harsh Trivedi, Aruna Balasubramanian, and Niranjan
  Balasubramanian. 2020.
\newblock {DeFormer}: Decomposing pre-trained {Transformers} for faster
  question answering.
\newblock In \emph{ACL}.

\bibitem[{Chen et~al.(2020)Chen, Frankle, Chang, Liu, Zhang, Wang, and
  Carbin}]{nips20-bert-lt}
Tianlong Chen, Jonathan Frankle, Shiyu Chang, Sijia Liu, Yang Zhang, Zhangyang
  Wang, and Michael Carbin. 2020.
\newblock The lottery ticket hypothesis for pre-trained {BERT} networks.
\newblock In \emph{NeurIPS}.

\bibitem[{{Colombo} and {Gao}(2020)}]{arxiv-our-binary}
Nicolo {Colombo} and Yang {Gao}. 2020.
\newblock \href {http://arxiv.org/abs/2009.05346} {Disentangling neural
  architectures and weights: A case study in supervised classification}.
\newblock \emph{arXiv:2009.05346}.

\bibitem[{Courbariaux et~al.(2015)Courbariaux, Bengio, and
  David}]{nips15-binnary-connect}
Matthieu Courbariaux, Yoshua Bengio, and Jean{-}Pierre David. 2015.
\newblock Binaryconnect: Training deep neural networks with binary weights
  during propagations.
\newblock In \emph{NeurIPS}.

\bibitem[{Devlin et~al.(2019)Devlin, Chang, Lee, and Toutanova}]{naacl19-bert}
Jacob Devlin, Ming{-}Wei Chang, Kenton Lee, and Kristina Toutanova. 2019.
\newblock {BERT:} pre-training of deep bidirectional transformers for language
  understanding.
\newblock In \emph{NAACL}.

\bibitem[{Fan et~al.(2020)Fan, Stock, Graham, Grave, Gribonval, Jegou, and
  Joulin}]{iclr20-quantization}
Angela Fan, Pierre Stock, Benjamin Graham, Edouard Grave, R{\' e}mi Gribonval,
  Herve Jegou, and Armand Joulin. 2020.
\newblock Training with quantization noise for extreme model compression.
\newblock In \emph{ICLR}.

\bibitem[{Frankle and Carbin(2019)}]{iclr19-lt}
Jonathan Frankle and Michael Carbin. 2019.
\newblock The lottery ticket hypothesis: Finding sparse, trainable neural
  networks.
\newblock In \emph{ICLR}.

\bibitem[{Frankle et~al.(2020{\natexlab{a}})Frankle, Dziugaite, Roy, and
  Carbin}]{icml20-bert-stablise}
Jonathan Frankle, Gintare~Karolina Dziugaite, Daniel~M. Roy, and Michael
  Carbin. 2020{\natexlab{a}}.
\newblock Linear mode connectivity and the lottery ticket hypothesis.
\newblock In \emph{ICML}.

\bibitem[{Frankle et~al.(2020{\natexlab{b}})Frankle, Dziugaite, Roy, and
  Carbin}]{arxiv-oneshot-bad}
Jonathan Frankle, Gintare~Karolina Dziugaite, Daniel~M. Roy, and Michael
  Carbin. 2020{\natexlab{b}}.
\newblock \href {http://arxiv.org/abs/2009.08576} {Pruning neural networks at
  initialization: Why are we missing the mark?}
\newblock \emph{arXiv:2009.08576}.

\bibitem[{{Gale} et~al.(2019){Gale}, {Elsen}, and {Hooker}}]{arxiv-mp-best}
Trevor {Gale}, Erich {Elsen}, and Sara {Hooker}. 2019.
\newblock \href {http://arxiv.org/abs/1902.09574} {The state of sparsity in
  deep neural networks}.
\newblock \emph{arXiv:1902.09574}.

\bibitem[{Ganesh et~al.(2020)Ganesh, Chen, Lou, Khan, Yang, Chen, Winslett,
  Sajjad, and Nakov}]{arxiv20-compress-bert-survey}
Prakhar Ganesh, Yao Chen, Xin Lou, Mohammad~Ali Khan, Yin Yang, Deming Chen,
  Marianne Winslett, Hassan Sajjad, and Preslav Nakov. 2020.
\newblock Compressing large-scale transformer-based models: A case study on
  {BERT}.
\newblock \emph{arXiv:2002.11985}.

\bibitem[{Gordon et~al.(2020)Gordon, Duh, and Andrews}]{rep4nlp-bert-transfer}
Mitchell~A. Gordon, Kevin Duh, and Nicholas Andrews. 2020.
\newblock \href {https://www.aclweb.org/anthology/2020.repl4nlp-1.18/}
  {Compressing {BERT:} studying the effects of weight pruning on transfer
  learning}.
\newblock In \emph{RepL4NLP@ACL}.

\bibitem[{He et~al.(2016)He, Zhang, Ren, and Sun}]{cvpr16-resnet}
Kaiming He, Xiangyu Zhang, Shaoqing Ren, and Jian Sun. 2016.
\newblock Deep residual learning for image recognition.
\newblock In \emph{CVPR}.

\bibitem[{Houlsby et~al.(2019)Houlsby, Giurgiu, Jastrzebski, Morrone,
  de~Laroussilhe, Gesmundo, Attariyan, and Gelly}]{icml19-adaptor}
Neil Houlsby, Andrei Giurgiu, Stanislaw Jastrzebski, Bruna Morrone, Quentin
  de~Laroussilhe, Andrea Gesmundo, Mona Attariyan, and Sylvain Gelly. 2019.
\newblock Parameter-efficient transfer learning for {NLP}.
\newblock In \emph{ICML}.

\bibitem[{Lan et~al.(2020)Lan, Chen, Goodman, Gimpel, Sharma, and
  Soricut}]{iclr20-albert}
Zhenzhong Lan, Mingda Chen, Sebastian Goodman, Kevin Gimpel, Piyush Sharma, and
  Radu Soricut. 2020.
\newblock {ALBERT:} {A} lite {BERT} for self-supervised learning of language
  representations.
\newblock In \emph{ICLR}.

\bibitem[{Li et~al.(2017)Li, De, Xu, Studer, Samet, and
  Goldstein}]{li2017training}
Hao Li, Soham De, Zheng Xu, Christoph Studer, Hanan Samet, and Tom Goldstein.
  2017.
\newblock Training quantized nets: A deeper understanding.
\newblock In \emph{NeurIPS}.

\bibitem[{Peters et~al.(2019)Peters, Ruder, and Smith}]{rep4nlp19-adapt}
Matthew~E. Peters, Sebastian Ruder, and Noah~A. Smith. 2019.
\newblock To tune or not to tune? adapting pretrained representations to
  diverse tasks.
\newblock In \emph{RepL4NLP}.

\bibitem[{Ramanujan et~al.(2020)Ramanujan, Wortsman, Kembhavi, Farhadi, and
  Rastegari}]{cvpr20-binary}
Vivek Ramanujan, Mitchell Wortsman, Aniruddha Kembhavi, Ali Farhadi, and
  Mohammad Rastegari. 2020.
\newblock What's hidden in a randomly weighted neural network?
\newblock In \emph{CVPR}.

\bibitem[{Rebuffi et~al.(2017)Rebuffi, Bilen, and Vedaldi}]{nips17-adapter}
Sylvestre-Alvise Rebuffi, Hakan Bilen, and Andrea Vedaldi. 2017.
\newblock Learning multiple visual domains with residual adapters.
\newblock In \emph{NeurIPS}.

\bibitem[{Savarese et~al.(2020)Savarese, Silva, and
  Maire}]{nips20-continuous-sparsification}
Pedro Savarese, Hugo Silva, and Michael Maire. 2020.
\newblock Winning the lottery with continuous sparsification.
\newblock In \emph{NeurIPS}.

\bibitem[{Shen et~al.(2020)Shen, Dong, Ye, Ma, Yao, Gholami, Mahoney, and
  Keutzer}]{aaai20-qbert}
Sheng Shen, Zhen Dong, Jiayu Ye, Linjian Ma, Zhewei Yao, Amir Gholami,
  Michael~W. Mahoney, and Kurt Keutzer. 2020.
\newblock {Q-BERT: Hessian Based Ultra Low Precision Quantization of BERT}.
\newblock In \emph{AAAI}.

\bibitem[{Wang et~al.(2019)Wang, Singh, Michael, Hill, Levy, and
  Bowman}]{iclr19-glue}
Alex Wang, Amanpreet Singh, Julian Michael, Felix Hill, Omer Levy, and Samuel
  Bowman. 2019.
\newblock {GLUE}: A multi-task benchmark and analysis platform for natural
  language understanding.
\newblock In \emph{ICLR}.

\bibitem[{Wortsman et~al.(2020)Wortsman, Ramanujan, Liu, Kembhavi, Rastegari,
  Yosinski, and Farhadi}]{nips20-supermask-sequence}
Mitchell Wortsman, Vivek Ramanujan, Rosanne Liu, Aniruddha Kembhavi, Mohammad
  Rastegari, Jason Yosinski, and Ali Farhadi. 2020.
\newblock Supermasks in superposition.
\newblock In \emph{NeurIPS}.

\bibitem[{Zhang et~al.(2020)Zhang, Wu, Katiyar, Weinberger, and
  Artzi}]{arxiv20-finetune-bert}
Tianyi Zhang, Felix Wu, Arzoo Katiyar, Kilian~Q. Weinberger, and Yoav Artzi.
  2020.
\newblock Revisiting few-sample {BERT} fine-tuning.
\newblock \emph{arXiv:2006.05987}.

\bibitem[{Zhou et~al.(2019)Zhou, Lan, Liu, and Yosinski}]{nips19-supermask}
Hattie Zhou, Janice Lan, Rosanne Liu, and Jason Yosinski. 2019.
\newblock Deconstructing lottery tickets: Zeros, signs, and the supermask.
\newblock In \emph{NeurIPS}.

\bibitem[{Zhu and Gupta(2018)}]{iclr18-mag-prune}
Michael Zhu and Suyog Gupta. 2018.
\newblock To prune, or not to prune: Exploring the efficacy of pruning for
  model compression.
\newblock In \emph{ICLR Workshop Track}.

\end{thebibliography}



\end{document}